\documentclass[conference]{IEEEtran}
\IEEEoverridecommandlockouts
\usepackage{cite}
\usepackage{amsmath,amssymb,amsfonts}
\usepackage{algorithmic}
\usepackage{graphicx}
\usepackage{textcomp}
\usepackage{xcolor}
\usepackage{hyperref}
\usepackage{multirow}
\usepackage{balance}

\def\BibTeX{{\rm B\kern-.05em{\sc i\kern-.025em b}\kern-.08em
    T\kern-.1667em\lower.7ex\hbox{E}\kern-.125emX}}
    
\begin{document}

\title{ROSAnnotator: A Web Application for ROSBag Data Analysis in Human-Robot Interaction
}

\author{
\IEEEauthorblockN{
% 1\textsuperscript{st} 
Yan Zhang}
\IEEEauthorblockA{
\textit{University of Melbourne}\\
Melbourne, Australia \\
yan.zhang.1@unimelb.edu.au
}
\and
\IEEEauthorblockN{
% 2\textsuperscript{nd} 
Haoqi Li}
\IEEEauthorblockA{
\textit{University of Melbourne}\\
Melbourne, Australia \\
haoqi2@student.unimelb.edu.au
}
\and
\IEEEauthorblockN{
% 3\textsuperscript{rd} 
Ramtin Tabatabaei}
\IEEEauthorblockA{
\textit{University of Melbourne}\\
Melbourne, Australia \\
stabatabaeim@\\student.unimelb.edu.au
}
\and
\IEEEauthorblockN{
% 4\textsuperscript{th} 
Wafa Johal}
\IEEEauthorblockA{
\textit{University of Melbourne}\\
Melbourne, Australia \\
wafa.johal@unimelb.edu.au
}
}

\maketitle

\begin{abstract}
Human-robot interaction (HRI) is an interdisciplinary field that utilises both quantitative and qualitative methods. While ROSBags, a file format within the Robot Operating System (ROS), offer an efficient means of collecting temporally synched multimodal data in empirical studies with real robots, there is a lack of tools specifically designed to integrate qualitative coding and analysis functions with ROSBags. To address this gap, we developed ROSAnnotator, a web-based application that incorporates a multimodal Large Language Model (LLM) to support both manual and automated annotation of ROSBag data. ROSAnnotator currently facilitates video, audio, and transcription annotations and provides an open interface for custom ROS messages and tools. By using ROSAnnotator, researchers can streamline the qualitative analysis process, create a more cohesive analysis pipeline, and quickly access statistical summaries of annotations, thereby enhancing the overall efficiency of HRI data analysis. \url{https://github.com/CHRI-Lab/ROSAnnotator}
\end{abstract}


\section{Introduction}
Human-robot interaction (HRI) is a field that acknowledges the importance of empirical study, which involves the collection of multimodal data to gain insights into human behaviour and interactions with robots. Given that many measurements cannot be simply quantified, qualitative methods are critical and typical research approaches in HRI~\cite{veling2021qualitative, bethel2010review}. Conducting qualitative analysis requires recording various types of data during experiments, including but not limited to video, audio, and robot actions. This process often involves using multiple devices for different purposes and necessitates careful time synchronization, which can be challenging. In HRI studies with a real robot, ROSBag offers an effective solution to streamline the recording and synchronization of such data.

ROSBag is a powerful tool within the Robot Operating System (ROS)~\cite{quigley2009ros} that allows for the recording of message data, offering researchers the flexibility to customise the types of data captured during a study. This data can be automatically synchronised and easily replayed for further analysis. Previous studies have advanced the development of ROS tools to support HRI studies. For instance, HRItk~\cite{lane2012hritk} is a toolkit designed to provide speech, gesture, and gaze recognition topics. Another study~\cite{mohamed2021ros} introduced the ROS4HRI framework, which is open-sourced and includes new message types specifically designed for HRI-related topics, along with a set of conventions and standard interfaces. These contributions enriched the types of data that can be collected using ROSBag. However, despite these advancements in data collection, the analysis of qualitative data remains a challenge.

Annotation, or coding, is an essential process for methods such as observation and interview, which are the two most frequently used approaches in HRI qualitative research~\cite{veling2021qualitative}. The annotation of observational data is selective and highly dependent on the study's purpose~\cite{papen2019participant}. Traditionally, annotation has been conducted manually, involving tasks like coding participants' utterances and emotions~\cite{prabhu2024dynamics}, often utilizing established tools such as ELAN~\cite{wittenburg2006elan} and ATLAS.ti~\cite{muhr1991atlas}. Moreover, to ensure objectivity and reliability, this process typically requires the involvement of multiple researchers and the assessment of inter-rater reliability~\cite{mchugh2012interrater}. Therefore, data analysis is often time-consuming and labour-intensive.

To fasten the qualitative pre-processing, recent work in human-computer interaction has been exploring ways to automate the coding of data. For instance, prior studies have successfully captured kinematic features from videos to interpret actions and gestures, thereby facilitating qualitative analysis~\cite{trujillo2019toward}. In addition, facial emotion recognition has been extensively investigated~\cite{canal2022survey}. With the advent of large language models (LLMs) and their advanced capabilities in processing text-based data, some researchers are now exploring their usage for deductive coding, achieving higher inter-rater reliability with experts by using a pre-defined codebook~\cite{xiao2023supporting}. However, these methods are not integrated and lack support for HRI researchers who are dealing with multimodal and robot logs packaged in ROSBags.

\begin{figure*}
    \centering
    \includegraphics[width=0.75\linewidth]{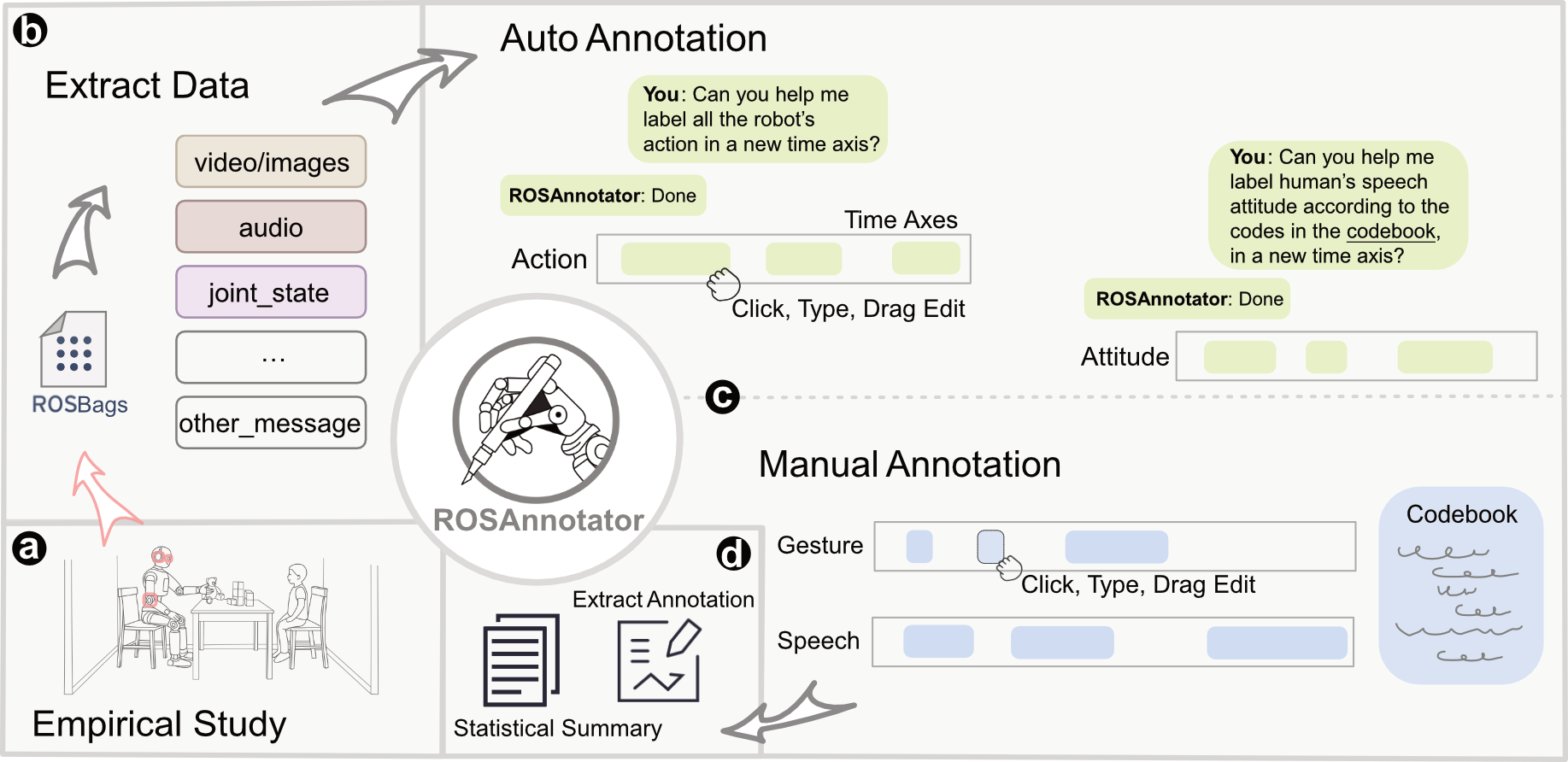}
    \caption{This figure illustrates the workflow of ROSAnnotation. (a) The process begins with users collecting ROSBags during an empirical study. (b) Once the ROSBags are imported into ROSAnnotator, multimodal data is extracted. (c) ROSAnnotator supports both automatic and manual annotation. For automatic annotation, users can provide instructions via a chatbox, and ROSAnnotator will perform the annotations accordingly. Users can then make manual adjustments if needed. For manual annotation, users can refer to the codebook, create multiple time axes, generate annotations, and modify the time intervals for annotations. (d) Finally, users can view a statistical summary of their qualitative analysis and extract the resulting annotations.}
    \label{fig:rosannotator}
\end{figure*}

Therefore, we present ROSAnnotator, a web application (WebApp) designed for HRI qualitative data analysis that is particularly suited for multimodal data. It supports both manual annotation and the multimodal Large Language Model (LLM)-facilitated annotation, which makes the analysis process more efficient. Additionally, ROSAnnotator supports cross-modal annotation, allowing users to annotate across different modalities, such as using audio data to inform video content annotation. The primary features (see \autoref{fig:rosannotator}) of ROSAnnotator include:
\begin{enumerate}
    \item Extracting messages from ROSBags and exporting annotation results.
    \item Displaying video, time axes for multiple annotation tiers, a multimodal LLM chatbox, and other toolbars within a web interface.
    \item Automatically transcribing audio and presenting it as annotations along a time axis.
    \item Allowing for a customised codebook and manual annotation of multimodal data on separate annotation tiers.
    \item Allowing for automated annotation, with codes generated on time axes by the multimodal LLM.
    \item Providing interfaces for integrating other customised messages and tools.
    \item Offering statistical summary for all annotations.
\end{enumerate}

\section{ROSAnnotator Functions}
\begin{figure}
    \centering
    \includegraphics[width=0.9\linewidth]{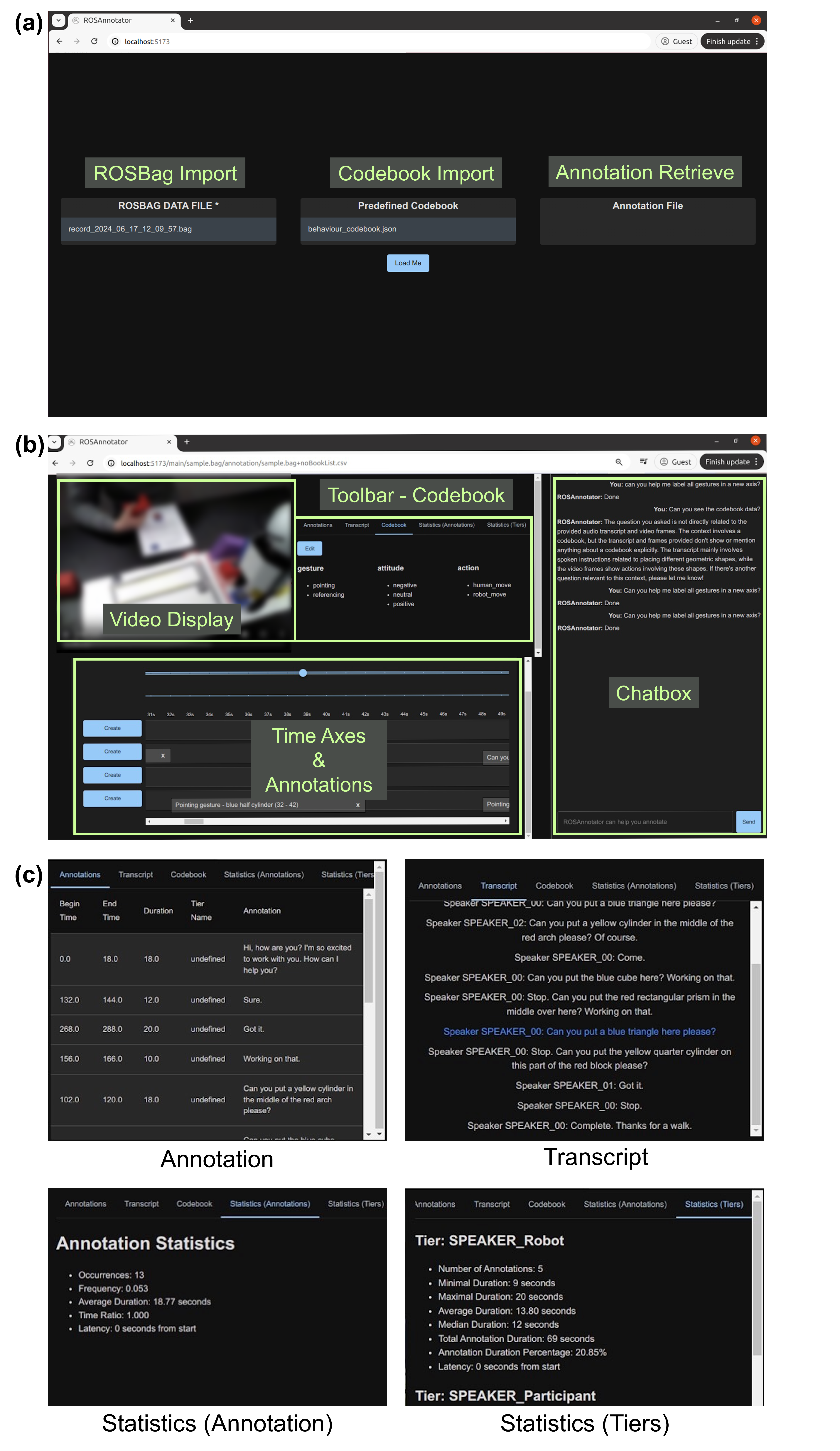}
    \caption{The interface of ROSAnnotator WebApp. (a) shows the import page, (b) shows the annotation page with the codebook function in the toolbar, (c) shows other functions in the toolbar}
    \label{fig:interface}
\end{figure}

The user journey begins with the process of data import (see \autoref{fig:interface}a).

\textbf{Data Import:} Users can select the desired ROSBag from a list to begin their work. It is also recommended that a pre-defined codebook in JSON format be uploaded to streamline the annotation process. Additionally, if audio transcription is required, users have the option to transcribe the audio simultaneously.

The main annotation interface is shown as \autoref{fig:interface}b.

\textbf{Data Visualisation:} The interface features a video player located in the top left corner, which synchronises with audio playback if audio extraction has been performed. In the top middle section of the interface, a toolbar *see \autoref{fig:interface}b and c) allows users to conveniently manage and edit their codebook, review annotations, access transcription data (which plays in tandem with the audio), and view a statistical summary of their annotations. The bottom section of the page displays the time axes, where users can create multiple tiers for various purposes, such as one tier for utterances, another for gestures, and another for emotions. When transcription data is available, utterances from each speaker are automatically organised into separate tiers, with each utterance displayed in its own tier and marked at the time interval during which it was spoken. On the right-hand side of the page, a chatbox is available for multimodal LLM-assisted annotations.

\textbf{Manual Annotation:} Users can create two types of annotation tiers: one that allows selection from a pre-defined codebook via a drop-down menu and another that enables free text input. Both types support direct editing and dragging along the time axes. Users can also define the start time, end time, duration, tier name, and the content of the annotation through the toolbar.

\textbf{Automated Annotation:} The embedded multimodal LLM serves to expedite the annotation process. Users can interact with the LLM via a chatbox, providing instructions, such as requesting assistance with annotations based on the transcription or video content. The LLM will then create a new tier and apply the corresponding annotations at the relevant time intervals. Additionally, users have the flexibility to edit the LLM-generated annotations either within the time axes or through the annotation toolbar. To address privacy concerns, a local algorithm has been implemented to detect and remove frames containing human faces before the video data is uploaded to the LLM.

\textbf{Statistical Summary:} The ROSAnnotator provides a detailed statistical summary for all the annotations, as well as for each individual tier. Key metrics for all tiers, including occurrences, frequency, average duration, time ratio, and annotation latency, are computed. For each tier, annotations are separately summarised, providing detailed statistics such as the total number of annotations, the minimum and maximum duration of a single annotation, the average and median duration, the total annotation duration, the percentage of time occupied by annotations, and the latency. The meanings of statistical metrics are summarised in \autoref{tb:statistic}.

\textbf{Data Export:} ROSAnnotator allows for the export of various types of data. Processed video, audio, and transcription files are saved within the designated data directory. If the pre-defined codebook has been modified within ROSAnnotator, it will be updated accordingly. All annotations, including their content, tier name, start time, and end time, can be exported to a CSV file for further analysis. Additionally, a statistical summary of the annotations can be exported if required.

\section{Installation}
As the ROSAnnotator is developed as a web application, it is platform-independent and thus not constrained by the user's operating system. Before initiating the installation process, it is essential to ensure that Docker~\cite{merkel2014docker} is installed on your machine. If Docker is not already installed, it can be downloaded and installed from the official Docker\footnote{https://www.docker.com/get-started/} website. Prior to launching the server, navigate to the root directory of the project and build the image, which has an approximate size of 20 GB. ROSAnnotator requires an OpenAI Key and a Hugging Face Access Token to enable certain functionalities. To configure these, a \texttt{.env} file must be created in the root directory of the backend. In this file, the two required environment variables, OPENAI\_API\_KEY\footnote{https://platform.openai.com/api-keys} and HUGGINGFACE\_AUTH\_TOKEN\footnote{https://huggingface.co/docs/hub/datasets-polars-auth}, should be manually added. Proper configuration of these credentials is essential for the application to operate as intended.

Both the frontend and backend are hosted locally within Docker containers. Once the server has been successfully initiated, the web application can be accessed through a web browser at http://localhost:5173/, which will provide users with the locally hosted interface of the application.

\begin{table*}[]
\centering
\caption{This table lists the meaning of the statistical metrics.}
\label{tb:statistic}
\resizebox{0.9\linewidth}{!}{
\begin{tabular}{lll}
\hline\hline
\multicolumn{1}{l}{}     & Metrics          & Meaning                                                           \\ \hline
\multirow{5}{*}{Overall} & Occurrences      & The total number of annotations.                                  \\
                         & Frequency        & The rate at which annotations appear over a given period of time. \\
                         & Average Duration & The average length of time an annotation lasts.                   \\
 &
  Time Ratio &
  The proportion of total annotation time relative to the observation period. \\
                         & Latency          & The time delay before the first annotation occurs.                \\ \hline
\multirow{8}{*}{Individual Tier} &
  Number of Annotations &
  The total number of annotations for this tier. \\
                         & Minimal Duration & The shortest duration of annotations for this tier.                             \\
                         & Maximal Duration & The longest duration of annotations for this tier.                              \\
                         & Average Duration & The average duration of annotations for this tier.                              \\
                         & Median Duration  & The median duration of annotations for this tier.                               \\
                         & Total Annotation Duration   & The sum of the duration of all annotations for this tier.                       \\
 &
  Annotation Duration Percentage &
  The proportion of the total observation time in this tier that is occupied by annotations. \\
                         & Latency          & Same as the overall one, but only for this tier.                  \\ \hline\hline
\end{tabular}
}
\end{table*}

\section{Code Architecture}
\begin{figure}
    \centering
    \includegraphics[width=1\linewidth]{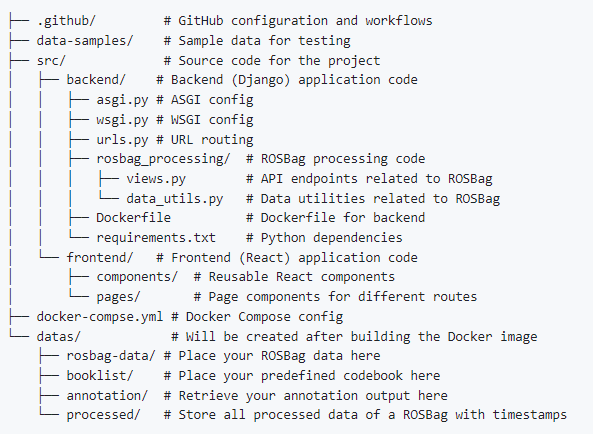}
    \caption{Code structure of the ROSAnnotator WebApp}
    \label{fig:structure}
\end{figure}

\autoref{fig:structure} shows the code architecture of the ROSAnnotator WebApp. The frontend is built using the React library~\cite{gackenheimer2015introduction}, while the backend, responsible for managing ROSBags and delivering data to the frontend, is implemented with the Django framework~\cite{forcier2008python}. Within the \texttt{rosbag\_processing} directory, the \texttt{data\_util.py} file contains functions for data extraction and allows users to define and extract messages. By default, the functions support video and audio extraction as well as audio transcription. However, users are required to modify the topic names in \texttt{data\_util.py} to align with the ones defined in their respective ROSBags. The \texttt{views.py} file in the same directory hosts the API endpoints, which currently support a manual annotation function and a privacy-secured LLM-based automatic annotation function. Users have the option to extend this file by adding other customised tools, such as gesture recognition. Furthermore, the file includes default prompts and actions for the LLM model, though users can modify these prompts or switch to a different model if necessary.

After building the Docker environment using the \texttt{docker-compose.yml} file, a \texttt{datas} folder will be automatically generated. Users can place their ROSBags in the \texttt{rosbag-data} directory and their predefined codebooks in the \texttt{booklist} directory. Annotations are saved in the \texttt{annotation} folder, and users can access their incomplete annotations from this location to resume their work. To enhance processing efficiency, each ROSBag is processed only once, with the processed data stored in the \texttt{processed} directory for direct access in subsequent operations.

\section{Usage Scenarios}
\textbf{Data analysis:} ROSAnnotator is a valuable tool for facilitating qualitative data analysis in HRI. For instance, in~\cite{wang2020see}, researchers examined participants' switch-hand behaviour, number of tool drops, and head movements while collaborating with a UR5 robot, using video footage captured by a camera. Similarly, in~\cite{kuyucu2024human}, video coders assessed recordings from a Pepper Robot, rating the level of warmth exhibited by participants during interactions. ROSAnnotator can streamline these types of measurements, enhancing the efficiency and accuracy of the analysis process.

\textbf{Construct dataset:} ROSAnnotator also supports the labelling and construction of datasets. For instance, in~\cite{chen2022human}, researchers published a multimodal dataset, recorded using ROSBags, that included camera video, joint states, and other sensor data from robots. Additionally, researchers developed a conversational dataset~\cite{jayagopi2013vernissage} comprising both video and audio data. By using ROSAnnotator to label these recordings, datasets can be better organized and more adaptable for various research purposes.

\section{Limitations and Future Works}
As ROSAnnotator is a newly developed web application with novel features, it currently faces several limitations. First, the automated annotation relies on a commercial API, incurring costs and limiting accessibility, while transcription poses privacy risks due to external audio processing. Future plans include hosting models locally to ensure data security and anonymising transcripts with generic labels (e.g., "Speaker 1") to further protect identities. Additionally, due to the constraints of the multimodal LLM's capabilities, the performance of the automated annotation is not robust. While users can define codebooks with detailed code lists for auto-annotation, and the system supports multi-tier and multimodal annotations to facilitate more subtle analyses, addressing nuanced HRI contexts remains a challenge for auto-annotations. Furthermore, the current version of ROSAnnotator primarily supports video (\textbackslash image\_raw topic) and audio (\textbackslash audio topic) as default data types and also enables the use of user-defined messages. While users can already parse customised messages from ROSBags, the integration of AI-based analysis capabilities is planned for future development. To address these limitations, we will open-source ROSAnnotator and encourage HRI researchers to contribute by integrating custom message types and tools.

\section{Conclusion}
Qualitative research is gaining a presence in the HRI research landscape, as illustrated by a new dedicated study track for qualitative methods. 
In this paper, we present ROSAnnotator, a web-based tool designed to support multimodal qualitative data analysis. Its primary advantages include: 1) the ability to directly extract messages from ROSBags and visualise data and time axes within a unified interface; 2) support for both manual annotations with codebooks and automated annotation through multimodal LLM; and (3) an open-source framework, which facilitates customised messages and tools for user-specific needs. ROSAnnotator aims to improve the efficiency of the qualitative data analysis process in empirical HRI studies and assist in labelling ROSBag data for the creation of datasets that can serve various research purposes.

\section*{Acknowledgment}
We gratefully acknowledge the support provided by the Melbourne Research Scholarship (University of Melbourne) and the Australian Research Council Discovery Early Career Research Award (Grant No. DE210100858). We thank Bowen Fan, Tianqi Wang, Guanqin Wang, Yujie Zheng, Yuchen Song, Yucheng Peng, and Abhishek Tummalapalli for their assistance during the early stages of this work.

%%%%%%%%%%%%%%%%%%%%% REFERENCES %%%%%%%%%%%%%%%%%%%%%
\balance

\bibliographystyle{ieeetr}
\bibliography{ref}

\end{document}